\documentclass[letterpaper, 10 pt, conference]{ieeeconf}  

\IEEEoverridecommandlockouts                              

\usepackage{hyperref}
\hypersetup{
    hidelinks
}
\usepackage{graphics} 
\usepackage{caption}
\usepackage{amsmath,amssymb}
\usepackage{bm}
\usepackage{mathtools}
\usepackage{graphicx}
\usepackage{amsmath,amssymb,bm}
\usepackage{booktabs,multirow,makecell}
\usepackage{graphicx,array,xcolor}

\title{\LARGE \bf
From Reasoning to Action: Consequence-Grounded Alignment \\
  for Vision-Language-Action Models
}

\author{
Yixue Zhang$^{1,2}$,
Wenke Xia$^{2,3}$,
Kun Wu$^{2}$,
Wenbo Yu$^{2,4}$,
Pei Ren$^{2},$\\[0.5em]
Junnan Nie$^{1}$,
Di Hu$^{3}$,
Zhengping Che$^{2,*}$,
Chang Liu$^{1,*}$
\thanks{
$^{1}$Peking University, Beijing, China.
}
\thanks{
$^{2}$Beijing Innovation Center of Humanoid Robotics,
Beijing, China.
}
\thanks{
$^{3}$Renmin University of China, Beijing, China.
}
\thanks{
$^{4}$Beijing Forestry University, Beijing, China.
}
\thanks{
$^{*}$Corresponding authors:
Chang Liu (\texttt{changliucoe@pku.edu.cn}) and
Zhengping Che (\texttt{z.che@x-humanoid.com}).
}
}

\begin{document}

\maketitle
\thispagestyle{empty}
\pagestyle{empty}

\begin{abstract}

Reasoning informs decision-making by connecting intended actions with their expected consequences. Building on this principle, recent vision-language-action models incorporate embodied chains of thought (ECoTs) as an intermediate stage in action generation. However, such guidance remains implicit, as actions are conditioned on the generated ECoTs rather than explicitly optimized to align with them.
To bridge this gap, we introduce the Consequence-grounded Reasoning-to-Action alignment method called CoRAct, an offline training framework that explicitly aligns reasoning and action through reward-guided optimization in action noise space.
Specifically, CoRAct formulates two physically grounded ECoT components as verifiable signals: movement specifies the intended motion, whereas waypoints encode the expected spatial consequence of execution. By jointly modeling ECoT reasoning, action generation, and action-conditioned future prediction, a unified Mixture-of-Transformers explicitly couples each action with its articulated intent and predicted consequence. For action candidates generated from sampled noise, CoRAct computes alignment rewards based on motion consistency with the described movement and agreement between predicted outcomes and the specified waypoints. These rewards drive group-relative optimization of the initial noise distribution, steering action generation toward consistency with the articulated intent and expected consequence.
Extensive experiments across RoboCasa, VLABench, and real-world manipulation tasks demonstrate consistent gains in task success and generalization over strong ECoT-based VLA baselines. Further analyses reveal improved reasoning–execution alignment and show that our method can ground externally provided ECoTs in executable behavior, supporting zero-shot adaptation to novel scenarios by leveraging knowledge from powerful VLMs.
Our project page and dataset are publicly available at
\href{https://onlyxingxingzyx.github.io/CoRAct/}{https://onlyxingxingzyx.github.io/CoRAct/}.
\end{abstract}

\section{INTRODUCTION}

Human decision-making often involves prospective reasoning that connects intended actions with their expected consequences. Such reasoning mediates between task intent and concrete behavior by relating action formulation to anticipated outcomes. 
Inspired by this capability, recent vision-language-action (VLA) models increasingly integrate intermediate reasoning into action generation in the form of embodied chains of thought (ECoTs)~\cite{zawalski25a,sunemma,zhao2025cot,huang2025motvla,shou2026halo}. By structuring action generation around textual plans, visual subgoals, spatial targets, and action-level intentions, ECoTs provide an intermediate interface between task understanding and physical execution~\cite{ma2026unifying,lu2026thinkingvla}. As the paradigm evolves, embodied reasoning is increasingly integrated into the perception–action pipeline, extending beyond purely textual deliberation toward representations more closely tied to physical control.


Prior efforts have improved VLA generalization
through diversity-driven task generation
\cite{zhang2026robogene}, visual data augmentation
\cite{wang2026roboaug}, and unified vision-motion
representation learning \cite{fan2026xr1}.
While these approaches enhance generalization
across diverse tasks, objects, and visual conditions,
adapting to unseen semantic constraints and novel
task compositions requires reasoning beyond
learned action patterns.
Embodied reasoning offers a complementary path
toward such out-of-distribution generalization
by connecting task intent with anticipated
physical consequences.
However, its potential remains limited if
the generated actions fail to faithfully
realize the articulated reasoning. Existing VLA models typically use ECoTs as conditioning context for action generation, leaving this correspondence to be learned implicitly from demonstrations rather than enforced through an explicit optimization objective~\cite{huang2025motvla,shou2026halo,kim2026roboalign,sun2026atomvla}. In embodied control, this gap has concrete consequences: a model may articulate an appropriate motion yet generate an action that deviates from it, or anticipate a plausible spatial outcome that execution fails to realize. These discrepancies reveal a missing link between coherent intermediate reasoning and physical behavior that faithfully reflects both the stated intent and the expected outcome.

\begin{figure}[t]
    \centering
	\includegraphics[width=\linewidth]{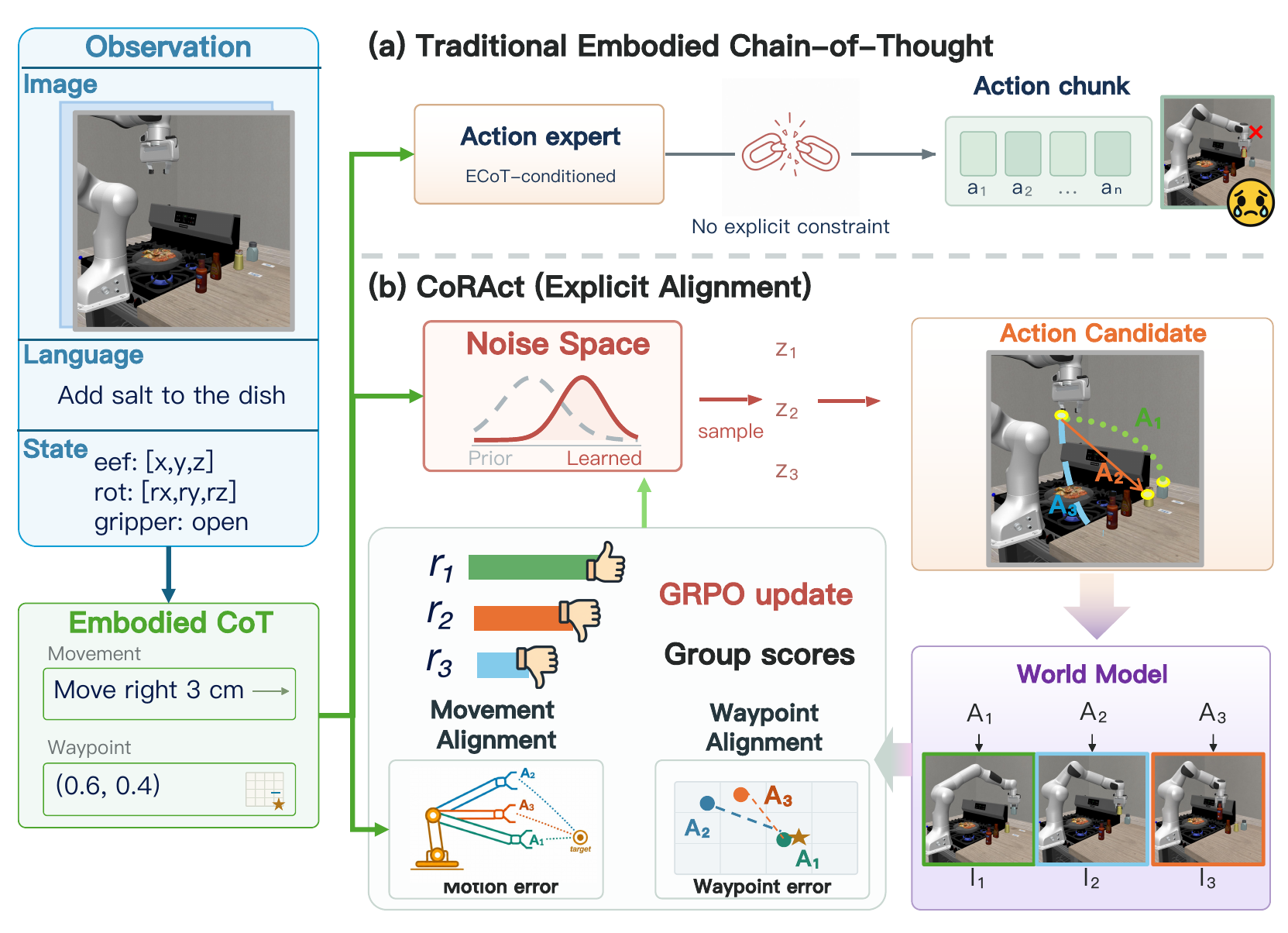}

  \caption{\textbf{Reasoning-to-action alignment paradigms.}
  (a) Direct CoT-to-action learning leaves reasoning--execution correspondence implicit.
  (b) CoRAct explicitly bridges reasoning and execution through consequence-grounded optimization of the action noise distribution, aligning generated actions with articulated intent and expected outcomes.}
    \label{fig:teaser}
\end{figure}

To bridge this gap, we introduce CoRAct, an offline training framework for consequence-grounded reasoning-to-action alignment. CoRAct explicitly optimizes the consistency between embodied reasoning and generated actions through reward-based learning in the action noise space. The key idea is to treat the motion intent and spatial expectations expressed in embodied reasoning as \textbf{verifiable constraints for action generation}. These signals provide concrete criteria for evaluating whether generated actions follow the articulated reasoning and whether their predicted consequences agree with the expected outcomes. By translating these consistency evaluations into alignment rewards, CoRAct establishes a direct connection between embodied reasoning and action optimization, enabling reasoning-aware improvement of robotic policies.

Specifically, CoRAct extracts two physically grounded signals from ECoT reasoning: movement for describing the intended motion and waypoints for representing the expected spatial consequence.
To connect these reasoning signals to action generation and future consequences, we build a unified Mixture-of-Transformers (MoT) architecture that jointly models ECoT reasoning, action generation, and action-conditioned future prediction. This joint formulation associates each generated action with both its articulated intent and predicted consequence, allowing the same action candidate to be evaluated against the reasoning that motivates it and the future outcome it is expected to produce. CoRAct therefore measures whether a generated action follows the described movement and whether its predicted consequence agrees with the specified waypoint, and converts these two forms of consistency into alignment rewards.
Building on these alignment signals, CoRAct optimizes action generation in the noise space. For each reasoning context, multiple action candidates are generated from sampled noise and evaluated for motion and consequence consistency. The resulting rewards drive group-relative optimization of a conditional noise predictor, guiding it toward noise samples that induce actions better aligned with both the intended motion and expected consequence. This improves the fidelity between ECoT reasoning and generated actions, strengthening reasoning-to-action alignment.

We evaluate CoRAct across RoboCasa, VLABench, and real-world manipulation tasks, demonstrating consistent improvements in task success and generalization over strong ECoT-based VLA baselines. Further analyses show stronger alignment between reasoning and execution, with the benefits extending beyond reasoning generated by the policy itself. In particular, CoRAct can ground ECoTs provided by powerful VLMs in executable actions, allowing their knowledge and decision-making capabilities to guide manipulation in novel scenarios. This capability enables zero-shot adaptation through external reasoning without additional policy training, extending ECoTs from an intermediate reasoning representation to a general interface for guiding physical behavior.

Overall, our main contributions are as follows:

\begin{itemize}
    \item We introduce CoRAct, a consequence-grounded alignment framework that explicitly makes consistency between embodied reasoning and physical execution.

    \item Building on joint training, we derive alignment rewards from motion consistency with movement descriptions and agreement between predicted outcomes and waypoints. These rewards drive group-relative policy optimization in action noise space, steering actions toward the articulated intent and expected consequences

    \item Experiments in simulation and the real world demonstrate higher task success and stronger generalization than strong ECoT-based VLA baselines. Further experiments show that improved reasoning–action alignment enables zero-shot adaptation to novel scenarios 
\end{itemize}

\section{RELATED WORKS}

\textbf{Embodied Chain-of-Thought Reasoning.} To improve generalization and interpretability in vision-language-action policies, embodied chain-of-thought (ECoT) introduces intermediate reasoning representations that connect semantic task understanding with grounded robot control. Early approaches instantiate reasoning in language: ECoT generates grounded textual traces~\cite{zawalski25a}, while EMMA-X augments them with look-ahead spatial guidance~\cite{sunemma}. Subsequent work broadens the reasoning substrate beyond text through representations including future visual subgoals~\cite{zhao2025cot}, reinforced visual planning~\cite{huang2026thinkact}, and coarse action trajectories~\cite{zhong2026acot}. Complementary work shifts reasoning from explicit generation into latent representations: ECoT-Lite studies how reasoning supervision shapes action-relevant representations~\cite{chen25a}, while LaRA-VLA~\cite{bai2026latent} and LaST$_0$~\cite{liu2026last} internalize multimodal and spatio-temporal CoTs in continuous latent states. ERVLA shows that ECoT supervision is not uniformly beneficial: high-level semantic traces alone yield limited gains, whereas action-grounded signals provide stronger guidance for control~\cite{sun2026revisiting}. Despite this progress, most methods induce correspondence between reasoning and action implicitly through conditioning supervision, without directly verifying whether policy-generated actions comply with the model's CoT. In contrast, CoRAct turns generated ECoTs from conditioning context into verifiable supervision, explicitly training the policy to translate articulated reasoning into physical execution.

\textbf{Consistency between Reasoning and Execution.} 
Consistency between articulated reasoning and physical execution is essential for reliable embodied decision-making. Prior work formulates this
  correspondence as explicit language–action alignment, using contrastive grounding and preference learning to align generated descriptions with visual
  observations and action trajectories~\cite{Wulff_2026_CVPR}. Beyond representation-level alignment, model-based approaches couple structured reasoning
  with low-level planning or action-conditioned future prediction~\cite{ma2026dvlm,tan2025latent,wu2025foresight}.
Feedback-driven policy optimization improves robotic
behavior through human-assisted action preferences
\cite{xia2025apo} and reliable value estimation
\cite{xia2026robovaluerl}.
Complementary approaches leverage external
semantic reasoning to guide fine-grained action
correction~\cite{xia2025phoenix}.
In contrast, CoRAct directly grounds a VLA's
embodied reasoning in action generation,
using explicit consistency rewards to align
articulated intent, executed motions,
and predicted consequences. Runtime verification selects among policy samples
  through simulated consequences or learned action verifiers~\cite{kwok2026scaling,wu2025you,kwok25a,dai2025rover}. These deployment-time methods require repeated
  policy sampling and external evaluation, often through an interactive simulator or a VLM-based verifier. To move this burden into training, recent
  work employs verifiable rewards, task-success signals, or learned critics to post-train reasoning VLAs~\cite{ye2025vla,yin2025deepthinkvla,wang2025alpamayo,foutter2026vision}. However, these objectives primarily assess textual reasoning, visual
  grounding, high-level trajectories, or episode-level success, rather than directly constraining correspondence between structured CoT fields and fine-
  grained action chunks. They do not jointly impose two sample-level constraints central to CoRAct: whether an action follows the generated CoT and
  whether its predicted consequence realizes the intended spatial outcome. CoRAct addresses both links by deriving complementary rewards for action
  adherence and consequence realization from each structured CoT, while distilling multi-sample verification into a lightweight noise policy that
  requires one action sample at deployment.

\begin{figure*}[t]
    \centering
	\includegraphics[width=\linewidth]{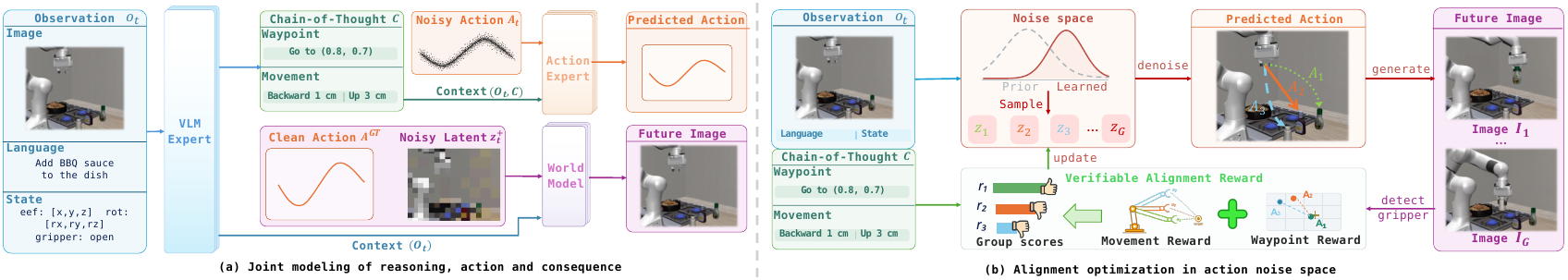}

\caption{
\textbf{Framework of CoRAct.}
(a) CoRAct jointly models ECoT reasoning, action generation, and action-conditioned future prediction in a unified Mixture-of-Transformers, coupling actions with intent and consequence.
(b) It then freezes the backbone and optimizes a conditional noise policy with movement and consequence rewards, steering action sampling toward better reasoning--execution alignment.
}

\vspace{-1em}
    \label{fig:teaser}
\end{figure*}

\section{Method}
  \label{sec:method}

CoRAct follows a two-stage training framework followed by lightweight deployment. In the first stage, we jointly model embodied chain-of-thought (ECoT) reasoning, action generation, and action-conditioned future prediction within a unified Mixture-of-Transformers (MoT) architecture, establishing the reasoning--action--consequence pathway described in Sec.~\ref{sec:joint-modeling}. In the second stage, we freeze the backbone and optimize a conditional noise policy with verifiable alignment rewards to explicitly improve consistency between articulated intent, generated actions, and predicted consequences, as detailed in Sec.~\ref{sec:grpo-training}. At inference, the learned alignment ability is retained to enable efficient single-sample action generation as described in Sec.~\ref{sec:inference}.


\subsection{Joint Modeling of Reasoning, Actions, and Consequences}
\label{sec:joint-modeling}

Consequence-grounded alignment requires connecting actions to articulated intent and evaluating whether their predicted consequences agree with the expected outcomes. We make these requirements explicit through two ECoT components, \emph{movement} and \emph{waypoints}. Both are constructed through rule-based annotation of demonstrations: movement is derived from relative end-effector poses and gripper actions, while waypoints are obtained by projecting the future gripper position into the camera view.

We jointly model ECoT reasoning, action generation, and future prediction within a unified Mixture-of-Transformers (MoT) backbone~\cite{liang2024mixture} comprising three specialized Transformer experts. Let $o=(I,\bm{s},\ell)$ denote the multimodal input, consisting of visual observations $I$, robot state $\bm{s}$, and language instruction $\ell$. We denote the structured ECoT by $C$, an action chunk of horizon $H$ by $\bm{A}=(\bm{a}_1,\ldots,\bm{a}_H)$, and the subsequent visual observation by $I^{+}$. The VLM expert processes the multimodal context and autoregressively predicts structured ECoT tokens, while the action and image generation experts model continuous action chunks and future visual latents through flow matching.  This design maintains separate parameters for reasoning, action generation, and future prediction while allowing controlled interaction among them through attention mask design. The backbone is trained through two streams with distinct sample formats and conditioning structures, as follows:

\paragraph{Reasoning-conditioned action generation}
The first stream jointly models ECoT and action generation as
\begin{equation}
    p_{\theta}(C,\bm{A}\mid o)
    =
    p_{\theta}(C\mid o)\,
    p_{\theta}(\bm{A}\mid o,C).
    \label{eq:reasoning-action-model}
\end{equation}
The VLM expert generates $C$ autoregressively using a token-level cross-entropy objective. Conditioned on $o$ and $C$, the action expert learns a conditional flow from action noise to the demonstrated action chunk.

Given a clean action chunk $\bm{A}^{\mathrm{gt}}$, we sample $\bm{\epsilon}_{a}\sim\mathcal{N}(\bm{0},\bm{I})$ and $t\sim\mathcal{U}(0,1)$, and construct the noised action state $\bm{A}_{t} =\alpha_t\bm{A}^{\mathrm{gt}}+\sigma_t\bm{\epsilon}_{a}$. The action expert is optimized with
\begin{equation}
    \mathcal{L}_{\mathrm{act}}
    =
    \mathbb{E}
    \left[
        \left\|
        v_{\theta}^{\mathrm{act}}
        (\bm{A}_{t},t\mid o,C)
        -
        \bm{u}_{t}^{\mathrm{act}}
        \right\|_2^2
    \right],
    \label{eq:action-flow-objective}
\end{equation}
where $\bm{u}_{t}^{\mathrm{act}}$ denotes the target velocity induced by the chosen action flow path. Through the attention mask, action tokens attend to the multimodal context and ECoT prefix, with bidirectional attention within the action block. This stream establishes an implicit correspondence between articulated reasoning and generated actions, which is explicitly strengthened in the subsequent alignment stage.

\paragraph{Action-conditioned future prediction}
The second stream models the visual consequence of executing an action. We encode the subsequent observation into the latent space of the VAE used by JanusFlow, $\bm{z}^{+}=\mathcal{E}_{\mathrm{VAE}}(I^{+})$, and train the image generation expert to predict this future latent conditioned on $o$ and the clean demonstrated action $\bm{A}^{\mathrm{gt}}$.

Given visual noise $\bm{\epsilon}_{I}\sim\mathcal{N}(\bm{0},\bm{I})$, we construct the noised future latent $\bm{z}^{+}_{t} =\alpha_t\bm{z}^{+}+\sigma_t\bm{\epsilon}_{I}$. The image generation expert is optimized with
\begin{equation}
    \mathcal{L}_{\mathrm{world}}
    =
    \mathbb{E}
    \left[
        \left\|
        v_{\theta}^{\mathrm{img}}
        (\bm{z}^{+}_{t},t\mid o,\bm{A}^{\mathrm{gt}})
        -
        \bm{u}_{t}^{\mathrm{img}}
        \right\|_2^2
    \right],
    \label{eq:world-flow-objective}
\end{equation}
where $\bm{u}_{t}^{\mathrm{img}}$ denotes the target velocity.

ECoT tokens are excluded from this stream, ensuring that future prediction depends only on the current observation and clean demonstrated action, thereby preventing the image generation expert from accessing the articulated waypoint. At inference, integrating the learned flow produces a predicted future latent $\hat{\bm{z}}^{+}$, which is decoded into a complete future image as $\hat{I}^{+} =\mathcal{D}_{\mathrm{VAE}}(\hat{\bm{z}}^{+})$. The decoded prediction is subsequently used for waypoint-grounded consequence evaluation.

\paragraph{Interleaved joint training}

At each optimization step, we sample the reasoning--action stream with probability $\eta$ and the action--world stream with probability $1-\eta$. Let $\mathcal{D}_{\mathrm{RA}}$ and $\mathcal{D}_{\mathrm{AW}}$ denote the corresponding sample distributions. The overall objective is
\begin{equation}
\begin{aligned}
    \mathcal{L}_{\mathrm{base}}
    ={}&
    \eta\,
    \mathbb{E}_{\mathcal{D}_{\mathrm{RA}}}
    \bigl[
        \mathcal{L}_{\mathrm{CoT}}
        + \mathcal{L}_{\mathrm{act}}
    \bigr]
    \\
    &+
    (1-\eta)\,
    \mathbb{E}_{\mathcal{D}_{\mathrm{AW}}}
    \bigl[
        \mathcal{L}_{\mathrm{world}}
    \bigr].
\end{aligned}
\label{eq:base-objective}
\end{equation}
Here, $\mathcal{L}_{\mathrm{CoT}}$ is the autoregressive cross-entropy loss.

This interleaved training equips the MoT backbone with reasoning-conditioned action generation and action-conditioned consequence prediction. These capabilities are subsequently used to evaluate whether generated actions follow the described movement and whether their predicted consequences agree with the specified waypoints.

\subsection{Alignment Optimization in Action Noise Space}
\label{sec:grpo-training}

The joint model in Sec.~\ref{sec:joint-modeling} establishes
reasoning-conditioned action generation and action-conditioned
consequence prediction.
However, base training induces correspondence between reasoning and
execution only through conditional generation, without explicitly
favoring action samples that more faithfully realize the articulated
intent and expected consequence.
We therefore freeze the MoT backbone and use alignment rewards to
optimize a lightweight conditional policy over action-flow
initializations, thereby explicitly improving sample-level
reasoning-to-action alignment.

\paragraph{Conditional noise policy}
During alignment training, each pair $(o,C^{\mathrm{gt}})$ is encoded
by the frozen MoT backbone into a context representation $\bm{h}$.
We replace the context-independent Gaussian prior used in base training
with a context-conditioned diagonal Gaussian policy~\cite{li2025gr,wagenmaker2025steering}:
\begin{equation}
    q_{\phi}(\bm{\epsilon}_{a}\mid\bm{h})
    =
    \mathcal{N}\!\left(
        \bm{\mu}_{\phi}(\bm{h}),
        \operatorname{diag}
        \bigl(\bm{\sigma}_{\phi}^{2}(\bm{h})\bigr)
    \right).
    \label{eq:noise-policy}
\end{equation}
For each context, we sample
$\{\bm{\epsilon}_{a,i}\}_{i=1}^{G}$ from
$q_{\phi_{\mathrm{old}}}(\cdot\mid\bm{h})$.
The frozen action expert then denoises each initialization for 10 steps  to produce an action candidate
$\bm{A}_{i}$.

\paragraph{Movement alignment reward}
We first assess whether each action candidate follows the movement
specified in $C^{\mathrm{gt}}$.
Let $\bm{m}^{k}$ denote the corresponding target for
$k\in\mathcal{K}=\{p,r,g\}$, representing translation, rotation, and
gripper command, and let $\mathcal{M}_{k}(\bm{A}_{i})$ extract the
corresponding endpoint motion from the action chunk.
After expressing the predicted and target motions under a common
coordinate convention, we compute the normalized error
$e_{i,k}
=\|\mathcal{M}_{k}(\bm{A}_{i})-\bm{m}^{k}\|_{2}/\tau_{k}$
and define
$R^{\mathrm{MOV}}_{i}
=-\sum_{k\in\mathcal{K}}\lambda_{k}e_{i,k}$,
where $\tau_k$ normalizes the scale of each motion component and
$\lambda_k$ controls its contribution.
This reward directly measures whether the sampled action realizes the
motion intent expressed in the provided ECoT.

\paragraph{Consequence alignment reward}
To verify the expected spatial consequence, we construct
waypoint annotations by projecting demonstrated future end-effector
positions into the corresponding camera images using calibrated
extrinsics and intrinsics.
These annotations are used to train a Grounding DINO-based waypoint
localizer $D_{\psi}$, which predicts the projected gripper waypoint
from an image.

For each action candidate $\bm{A}_{i}$, the image generation
expert predicts a future visual latent conditioned on
$(o,\bm{A}_{i})$, which is decoded by the VAE decoder into a future
image $\widehat{I}^{+}_{i}$.
The frozen waypoint localizer then extracts
$\widehat{\bm{u}}_{i}
=D_{\psi}(\widehat{I}^{+}_{i})$.
Let $\bm{u}^{\mathrm{ECoT}}$ denote the waypoint specified in
$C^{\mathrm{gt}}$, expressed in the same image coordinates.
We compute the waypoint error
$d_{u,i}
=\|\widehat{\bm{u}}_{i}-\bm{u}^{\mathrm{ECoT}}\|_{2}$
and define the consequence reward as
$R^{\mathrm{vis}}_{i}
=\exp(-d_{u,i}/\tau_{u})$,
where $\tau_u$ controls sensitivity to spatial prediction error.

Together, the movement and consequence terms provide complementary signals for candidate-level alignment:
\begin{equation}
    R_{i}
    =
    R^{\mathrm{MOV}}_{i}
    +
    \lambda_{\mathrm{vis}}R^{\mathrm{vis}}_{i}.
    \label{eq:dual-reward}
\end{equation}
The resulting reward favors actions that follow the movement specified
in the ECoT while producing predicted consequences consistent with its
waypoint.

\paragraph{Group-relative noise-policy optimization}
For candidates generated under the same observation and ECoT, the
combined rewards provide a relative measure of alignment quality.
We use the group mean as a critic-free baseline and define the
advantage of candidate $i$ as
$\widehat{A}_{i}
=R_{i}-\frac{1}{G}\sum_{j=1}^{G}R_{j}$.
We omit variance normalization to preserve the scale of reward
differences and avoid amplifying noise when the within-group variance
is small~\cite{shao2024deepseekmath}.

For each sampled initialization, we define the importance ratio
$\rho_{i}
=q_{\phi}(\bm{\epsilon}_{a,i}\mid\bm{h})/
q_{\phi_{\mathrm{old}}}(\bm{\epsilon}_{a,i}\mid\bm{h})$
and its clipped counterpart
$\bar{\rho}_{i}
=\operatorname{clip}
(\rho_{i},1-\varepsilon_{\mathrm{clip}},
1+\varepsilon_{\mathrm{clip}})$.
We optimize the conditional noise policy by maximizing
\begin{equation}
\begin{aligned}
    \mathcal{J}(\phi)
    =
    \mathbb{E}\Bigg[
        &\frac{1}{G}\sum_{i=1}^{G}
        \min\!\left(
            \rho_{i}\widehat{A}_{i},
            \bar{\rho}_{i}\widehat{A}_{i}
        \right)
        \\
        &+
        \beta\,
        \mathcal{H}
        \bigl(q_{\phi}(\cdot\mid\bm{h})\bigr)
    \Bigg],
    \label{eq:grpo-objective}
\end{aligned}
\end{equation}
where expectation is taken over offline contexts and
sampled candidate groups.
The clipping threshold $\varepsilon_{\mathrm{clip}}$ constrains the
policy update, while $\beta$ controls entropy regularization.

Only the conditional noise policy is updated, while the MoT backbone
and waypoint localizer remain frozen.
Alignment rewards reweight sampled action-flow initializations without
backpropagating through action generation, future prediction, or
waypoint localization, thereby favoring initializations that yield
better-aligned actions.

\subsection{Aligned Action Generation at Inference}
\label{sec:inference}
At inference, the frozen VLM expert first autoregressively generates an ECoT $C$ from the observation $o$, and the MoT backbone encodes $(o,C)$ into the context representation $\bm{h}$. The learned noise policy transfers the reasoning--action alignment preferences acquired from candidate-level verification directly into the action sampling distribution. Specifically, we sample a single  initialization $\bm{\epsilon}_{a}\sim q_{\phi}(\cdot\mid\bm{h})$, which is denoised by the frozen action expert for 10 steps conditioned on $(o,C)$ to produce an action chunk $\bm{A}$. In this way, CoRAct steers execution actions toward the articulated reasoning at the sampling stage to enable efficient deployment.

\begin{figure*}[!t]
    \centering

    \includegraphics[width=0.95\textwidth]{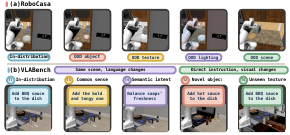}

    \vspace{-1mm}

    \captionof{figure}{
Generalization evaluation settings of RoboCasa and VLABench.
    }
    \label{fig:exp_setting}


\captionof{table}{
    In-distribution and generalization results on VLABench and RoboCasa.
    CoRAct consistently outperforms prior methods across these settings,
    benefiting from strong VLM generalization and improved reasoning--action consistency.
}
\label{tab:generalization}

{%
\scriptsize
\renewcommand{\arraystretch}{1.18}

\begin{tabular*}{\textwidth}{
    @{\extracolsep{\fill}}
    l
    *{5}{c}
    @{\hspace{6pt}}
    *{5}{c}
    @{}
}
    \toprule

    \multirow{2}{*}{\textbf{Method}}
    & \multicolumn{5}{c}{\textbf{VLABench}}
    & \multicolumn{5}{c}{\textbf{RoboCasa Atomic}} \\
    \cmidrule(lr){2-6}
    \cmidrule(lr){7-11}

    & \makecell{T1\\ID}
    & \makecell{T2\\Cat.}
    & \makecell{T3\\C.S.}
    & \makecell{T4\\Sem.}
    & \makecell{T6\\Tex.}
    & \makecell{In-\\Dis.}
    & \makecell{OOD\\Obj.}
    & \makecell{OOD\\Tex.}
    & \makecell{OOD\\Light.}
    & \makecell{OOD\\Scen.} \\
    \midrule

    $\pi_{0.5}$
    & 40.6 & 22.6 & 18.0 & 16.1 & 25.6
    & 72.6 & 67.8 & 59.6 & 64.7 & 51.6 \\

    $\pi_{0}$
    & 47.0 & 21.2 & 29.1 & 17.3 & \textbf{32.2}
    & 63.6 & 55.1 & 54.7 & 59.0 & 51.0 \\

    CoVer-VLA~\cite{kwok2026scaling}
    & 40.9 & 23.4 & 32.6 & 35.4 & 26.8
    & 65.1 & 58.4 & 56.6 & 60.1 & 52.1 \\

    $\mathrm{LaST}_{0}$
    & 40.9 & 21.1 & \textbf{34.4} & 35.0 & 26.6
    & 69.0 & 63.7 & 63.7 & 66.5 & 55.5 \\

    \midrule

    \textbf{CoRAct}
    & \textbf{51.8}
    & \textbf{24.3}
    & \textbf{34.4}
    & \textbf{36.2}
    & 30.0
    & \textbf{75.5}
    & \textbf{70.3}
    & \textbf{65.9}
    & \textbf{66.9}
    & \textbf{60.0} \\

    \bottomrule
\end{tabular*}
}%

    \vspace{-1em}
\end{figure*}

\providecommand{\TBD}{\textcolor{gray}{--}}
\providecommand{\best}[1]{\textbf{#1}}
\providecommand{\second}[1]{\underline{#1}}
\providecommand{\yes}{\ensuremath{\checkmark}}
\providecommand{\no}{\ensuremath{\times}}
\providecommand{\placeholderpanel}[3]{%
    \fbox{%
        \begin{minipage}[c][#2][c]{#1}
            \centering
            \footnotesize\bfseries #3\\[2pt]
            \scriptsize\normalfont Replace with final visualization
        \end{minipage}%
    }%
}
\providecommand{\maybegraphic}[4]{%
    \IfFileExists{#1}%
        {\includegraphics[width=#2,height=#3,keepaspectratio]{#1}}%
        {\placeholderpanel{#2}{#3}{#4}}%
}
\section{Experiments}
\label{sec:experiments}

We evaluate CoRAct from both task-level and mechanism-level perspectives. Experiments are organized to establish not only whether CoRAct improves robotic manipulation, but also whether the gains arise from stronger
reasoning-action correspondence. In the experiments, we attempt to address the following questions:

\begin{itemize}
    \item \textbf{Q1: Overall effectiveness.} How does CoRAct performs in in-distribution and out-of-distribution settings? Section~\ref{exp:comparison}.

    \item \textbf{Q2: Source of improvement.} Which components of CoRAct are responsible for
    the gains? Section~\ref{exp:ablation}.

    \item \textbf{Q3: Language following.} Does CoRAct improve
    compliance with fine-grained constraints on robotic action? Section~\ref{exp:language-following}.

    \item \textbf{Q4: Closed-loop self-correction.} Can the resulting policy
    recover from execution errors through subsequent observation-conditioned
    replanning? Section~\ref{exp:self-correction}.

    \item \textbf{Q5: Real-world transfer.} Can CoRAct ensure reliability in real-world scenarios?
    Section~\ref{exp:real-world}.
\end{itemize}

\subsection{Experimental Setup}
\label{exp:setup}

We instantiate our policy from JanusFlow-1.3B~\cite{ma2025janusflow}, leveraging its unified multimodal understanding and generation capability for structured CoT reasoning and future prediction, and augment the backbone with an action-specialized Transformer expert following a Mixture-of-Transformers design. During training, the model autoregressively generates a structured numeric CoT containing translational motion, rotation, gripper state, and a future gripper waypoint. Conditioned on the generated CoT, the action expert predicts an 8-step chunk of 7-D Cartesian actions via conditional flow matching, while the world-model branch predicts a future third-person observation in the latent space of an SDXL VAE conditioned on the action chunk~\cite{podell2024sdxl}. To prevent shortcut learning from the CoT, we mask CoT information from the world-model branch. For GRPO post-training, we freeze the base model and optimize only a 512-dimensional prefix-conditioned noise policy that controls the initialization of the frozen action flow. For each observation, we sample $G=8$ candidate action chunks and decode each using 10 flow integration steps, with candidates ranked by CoT--action consistency and action-conditioned visual waypoint alignment. To avoid artificially small action deviations when applying GRPO to observations already seen during backbone training, we split each dataset into 90\%/5\%/5\% subsets for backbone training, validation, and testing. We conduct GRPO post-training exclusively on the validation split and evaluate it on the held-out test split, ensuring a fair evaluation for subsequent quantitative analyses. For both RoboCasa and VLABench, we fine-tune the full backbone for 10 epochs in bfloat16 using AdamW with a learning rate of $2\times10^{-4}$, cosine decay, and weight decay of $0.01$. GRPO uses task-balanced sampling and is trained for 2,500 updates with a learning rate of $10^{-5}$. For evaluation, we test 17 RoboCasa tasks with 30 trials per task and 10 VLABench tasks with 50 trials per task, and report task success rate.

\subsection{Comparison Results}
\label{exp:comparison}

 CoRAct explicitly aligns structured embodied reasoning with downstream action generation, encouraging the policy to execute actions consistent with its inferred intent rather than merely producing plausible CoTs. We compare against representative VLA baselines, including $\pi_0$, $\pi_{0.5}$, CoVer-VLA~\cite{kwok2026scaling}, which explicitly supervises CoT--action correspondence through contrastive learning, and $\mathrm{LaST}_0$~\cite{liu2026last}, which uses latent CoT representations to implicitly guide action generation. We evaluate both in-distribution performance and generalization on RoboCasa~\cite{nasiriany2024robocasa} and VLABench~\cite{zhang2025vlabench}, as demonstrated in Figure~\ref{fig:exp_setting}.

As shown in Table~\ref{tab:generalization}, CoRAct consistently achieves the strongest overall performance across both benchmarks and their generalization settings, demonstrating the benefit of explicitly enforcing reasoning--action consistency over direct action learning or implicit reasoning guidance. Notably, RoboCasa emphasizes fine-grained manipulation across diverse atomic skills, while VLABench places greater emphasis on common-sense understanding, semantic reasoning, and visual generalization. CoRAct improves performance across both regimes, indicating that consequence-grounded reasoning--action alignment benefits both precise skill execution and higher-level embodied reasoning.

\subsection{Ablation and Quantitative Analysis}
\label{exp:ablation}

We first ablate the two ECoT components and their corresponding alignment objectives to examine where the performance gains of CoRAct originate. As shown in Table~\ref{tab:ablation}, introducing Movement CoT alone substantially improves success from 39.2\% to 44.3\% on VLABench and from 68.6\% to 72.9\% on RoboCasa, indicating that explicit motion-level reasoning already provides useful guidance for action generation. Adding Waypoint CoT further improves performance to 47.5\% and 73.5\%, respectively, with a larger gain on VLABench, where spatial and semantic reasoning plays a more prominent role.

Importantly, reasoning supervision alone does not fully realize these benefits. Explicitly aligning Movement CoT with the generated action further increases success to 49.3\% on VLABench and 74.6\% on RoboCasa. Adding waypoint--consequence alignment yields the full CoRAct model, reaching 51.8\% and 75.5\%, respectively. Overall, the complete model improves over the variant without CoT reasoning or alignment by 12.6\% and 6.9\% on VLABench and RoboCasa. These consistent incremental gains show that the improvement comes not only from introducing structured ECoT, but also from explicitly grounding the generated reasoning in actions and their predicted consequences.

\begin{table}[t]
    \centering
    \caption{
        Ablation results of CoT components and alignment objectives.
    }
    \label{tab:ablation}
    \vspace{2pt}
    \scriptsize
    \setlength{\tabcolsep}{4.0pt}
    \renewcommand{\arraystretch}{1.15}
    \begin{tabular}{@{}cccccc@{}}
        \toprule
        \multicolumn{2}{c}{CoT Components}
        & \multicolumn{2}{c}{Alignment Objectives}
        & \multicolumn{2}{c}{Success Rate (\%) $\uparrow$} \\
        \cmidrule(lr){1-2}
        \cmidrule(lr){3-4}
        \cmidrule(lr){5-6}
        \makecell{Move.\\CoT}
        & \makecell{Waypt.\\CoT}
        & \makecell{Move.\\Align.}
        & \makecell{Waypt.\\Align.}
        & VLABench
        & RoboCasa \\
        \midrule
        & & & &
        39.2\% & 68.6\% \\

        \checkmark & & & &
        44.3\% & 72.9\% \\

        \checkmark & \checkmark & & &
        47.5\% & 73.5\% \\

        \checkmark & \checkmark & \checkmark & &
        49.3\% & 74.6\% \\

        \midrule
        \checkmark & \checkmark & \checkmark & \checkmark
        & \textbf{51.8\%} & \textbf{75.5\%} \\
        \bottomrule
    \end{tabular}
\end{table}

\begin{table}[t]
    \centering
    \caption{
        Ablation of movement and waypoint alignment on quantitative reasoning--execution consistency metrics.
    }
    \label{tab:alignment-ablation}
    \vspace{2pt}
    \scriptsize
    \setlength{\tabcolsep}{5.0pt}
    \renewcommand{\arraystretch}{1.15}
    \begin{tabular}{@{}cccccc@{}}
        \toprule
        \multicolumn{2}{c}{Alignment}
        & \multicolumn{4}{c}{Consistency Metrics} \\
        \cmidrule(lr){1-2}
        \cmidrule(lr){3-6}

        \makecell{Move.\\Align.}
        & \makecell{Waypt.\\Align.}
        & \makecell{Pos.\\cm $\downarrow$}
        & \makecell{Rot.\\deg $\downarrow$}
        & \makecell{Grip.\\\% $\uparrow$}
        & \makecell{WP.\\px $\downarrow$} \\
        \midrule

        &
        & 0.401 & 0.534 & 69.789 & 5.998 \\

        \checkmark &
        & 0.209 & 0.478 & 87.661 & 2.904 \\

        &
        \checkmark
        & 0.232 & 0.494 & 85.160 & 2.706 \\

        \midrule
        \checkmark &
        \checkmark
        & \textbf{0.171}
        & \textbf{0.420}
        & \textbf{91.800}
        & \textbf{0.542} \\

        \bottomrule
    \end{tabular}
    \vspace{-1em}
\end{table}


\subsection{Fine-Grained Language Following}
\label{exp:language-following}

To directly evaluate reasoning--execution consistency, we measure four fine-grained metrics in Table~\ref{tab:alignment-ablation}. \textbf{Pos.} and \textbf{Rot.} measure the translational and rotational errors between the generated action and the movement specified by the ECoT, \textbf{Grip.} measures gripper-state agreement, and \textbf{WP.} measures the image-space distance between the predicted action consequence and the target waypoint. Both alignment objectives consistently improve these metrics, while combining them achieves the best results across all dimensions, reducing position error from 0.401 to 0.171~cm and waypoint error from 5.998 to 0.542~px, while increasing gripper agreement from 69.8\% to 91.8\%. These results show that CoRAct more faithfully translates fine-grained embodied reasoning into executable actions.

\subsection{Closed-Loop Self-Correction}
\label{exp:self-correction}

\begin{figure}[t]
    \centering
    \includegraphics[width=\linewidth]{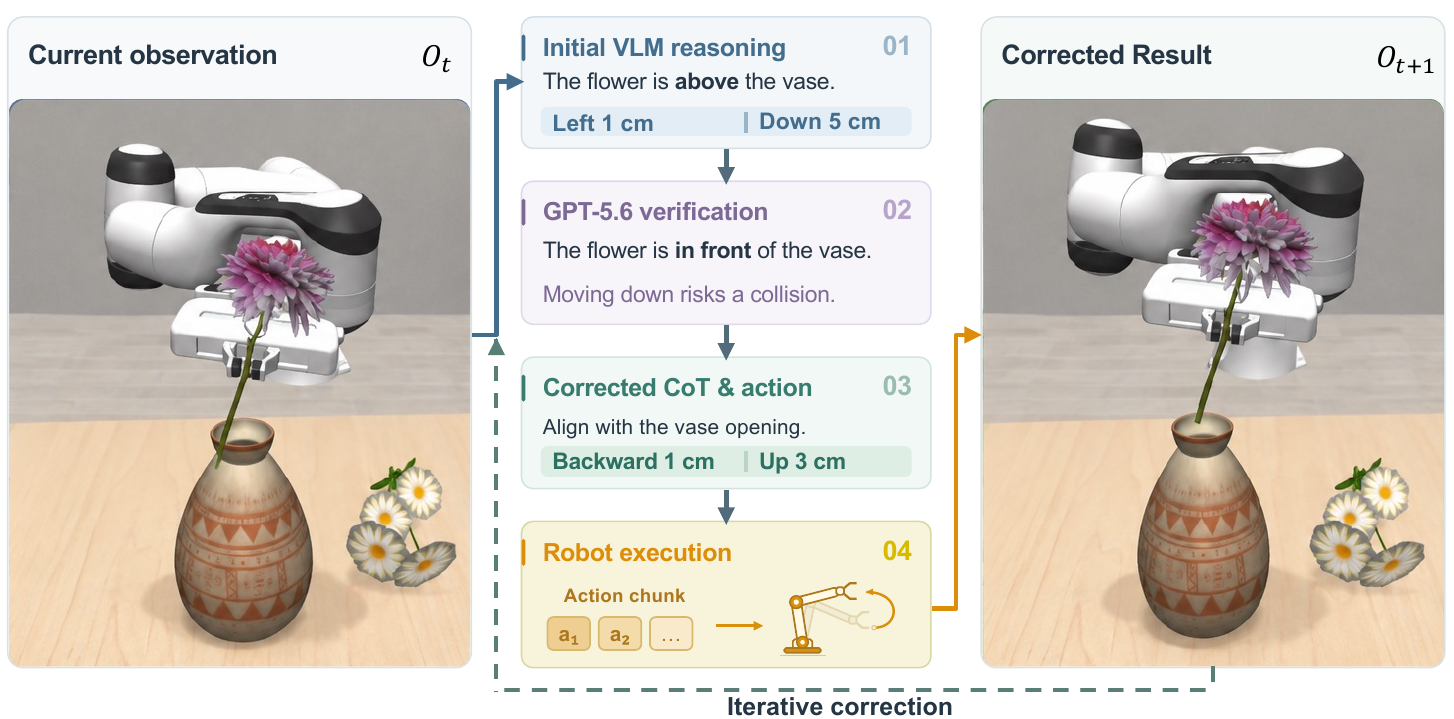}
    
    \caption{Illustration of our agent-guided correction pipeline, where GPT-5.6 Sol corrects the CoT generated by the VLM. }
    \label{fig:gpt_setting}
\end{figure}

CoRAct improves reasoning--action alignment by optimizing the action noise distribution. However, since post-training is performed only on expert trajectories, the policy has limited exposure to failure recovery and self-correction. Although Sec.~\ref{exp:comparison} demonstrates strong generalization to various unseen scenarios, failure states still induce unreliable ECoTs that cannot effectively guide recovery. To address this limitation, we introduce an external agent as a lightweight closed-loop correction module. As illustrated in Fig.~\ref{fig:gpt_setting}, when the policy produces incorrect ECoTs, the agent revises the reasoning and feeds the corrected ECoT back to CoRAct for action generation. This enables the policy to recover without additional training. In this setting, we only use movement as guided CoT for policy generation, quantitative results in Table~\ref{tab:self-correction} show that agent-guided ECoT correction consistently improves recovery performance, highlighting compatibility of CoRAct with stronger agentic reasoning.

\begin{table}[t]
    \centering
    \caption{
        Closed-loop self-correction on representative VLABench tasks.
        We report task success rates (\%).
    }
    \label{tab:self-correction}
    \vspace{2pt}
    \scriptsize
    \setlength{\tabcolsep}{7pt}
    \renewcommand{\arraystretch}{1.15}

    \begin{tabular}{@{}cccc@{}}
        \toprule
        \multicolumn{2}{c}{Components}
        & \multicolumn{2}{c}{Success Rate (\%) $\uparrow$} \\
        \cmidrule(lr){1-2}
        \cmidrule(lr){3-4}

        R--A Align.
        & Agent Corr.
        & Insert Flower
        & Chem. Tube \\
        \midrule

        &
        & 24.0 & 22.0 \\

        \checkmark &
        & 26.0 & 30.0 \\

        &
        \checkmark
        & 52.0 & 40.0 \\

        \midrule

        \checkmark &
        \checkmark
        & \textbf{66.0}
        & \textbf{52.0} \\

        \bottomrule
    \end{tabular}
    \vspace{-1em}
\end{table}

\subsection{Real-Robot Experiments}
\label{exp:real-world}

\begin{figure}[t]
    \centering
    \includegraphics[width=\linewidth]{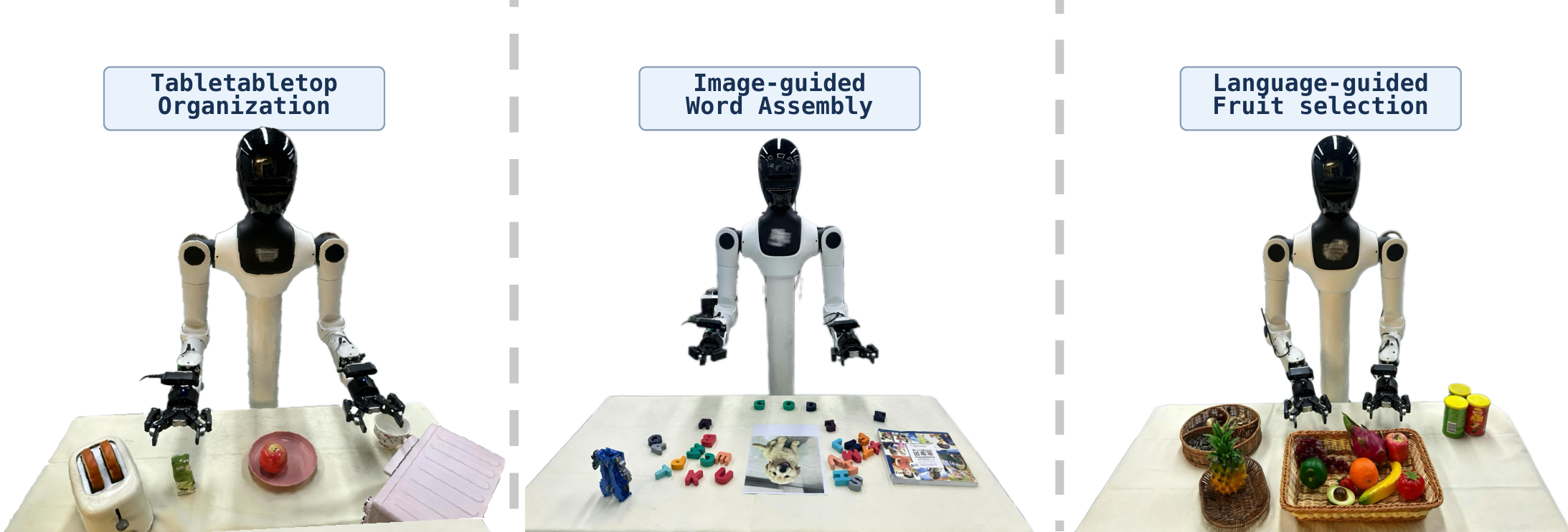}
    \caption{Real-world experimental setup.}
    \label{fig:real}
    \vspace{-1em}
\end{figure}

In real-world experiments, we evaluate three tasks for 30 trials: tabletop organization, image-guided word assembly, and language-guided fruit selection, covering long-horizon manipulation and semantic understanding capabilities . As illustrated in Figure~\ref{fig:real}, the robot needs to remove the apple from the plate and place the juice into it in tabletop organization task. In image-guided word assembly, given a reference image, the robot must select the corresponding letters from a set of 26 letters and arrange them to form the word. In fruit selection, the robot is instructed to identify and pick three specified types of fruit from a fruit bowl.

Due to constraints of master–slave teleoperation, we adapt the ECoT representation for the joint-space real-robot setting. Specifically, the \emph{movement} component predicts the movement of each robot joint, while the \emph{waypoint} component predicts projection of robot flange center onto the top camera view. To reduce the latency associated with autoregressive CoT generation, we attach dedicated prediction heads for joint movements and waypoints. During online inference, their outputs are assembled into a structured CoT using predefined rules to guide subsequent action generation. As shown in Table~\ref{tab:real-robot}, our method consistently improves performance across all three tasks, demonstrating stronger reasoning–execution alignment in long-horizon and semantics-intensive real-world manipulation.

\begin{table}[t]
    \centering
    \caption{
        Real-world evaluation on three long-horizon and semantic-intensive manipulation tasks.
    }
    \label{tab:real-robot}
    \vspace{2pt}
    \scriptsize
    \setlength{\tabcolsep}{4.5pt}
    \renewcommand{\arraystretch}{1.15}

    \begin{tabular}{@{}lccc@{}}
        \toprule
        Method
        & \makecell{Tabletop\\organization
        }
        & \makecell{Image-guided \\ word assembly}
        & \makecell{Language-guided\\fruit selection} \\
        \midrule

        $\pi_{0.5}$
        & 33.3 & 26.6 & 56.6 \\

        $\mathrm{LaST}_{0}$
        & 40.0  & 33.3 & 50.0 \\

        CoVer-VLA
        & 33.3 & 26.6 & 50.0 \\

        \midrule
        \textbf{CoRAct}
        & \textbf{56.6}
        & \textbf{46.6}
        & \textbf{63.3} \\

        \bottomrule
    \end{tabular}
    \vspace{-1em}
\end{table}

\section{CONCLUSION}

In this work, we presented CoRAct, an offline framework for explicitly aligning embodied reasoning with physical execution. Rather than treating ECoTs only as conditioning context for action generation, CoRAct formulates movement intent and waypoint consequences as verifiable signals for reasoning--action consistency optimization. By jointly modeling reasoning, actions, and action-conditioned future predictions, CoRAct derives consequence-grounded alignment rewards and uses them to optimize action generation in the noise space through group-relative learning. Experiments on RoboCasa, VLABench, and real-world manipulation demonstrate consistent improvements in task success and generalization, together with stronger agreement between generated reasoning and executed behavior. Moreover, CoRAct can ground externally provided ECoTs into executable actions, enabling knowledge from powerful VLMs to support zero-shot adaptation to novel scenarios. These results suggest that embodied reasoning can serve not only as an intermediate representation for action prediction, but also as a verifiable interface for explicitly optimizing physical behavior.

\textbf{Limitations and future work.}
Our current framework primarily focuses on reasoning signals that can be directly grounded in action execution. Extending explicit alignment to higher-level semantic reasoning, such as subgoals and long-horizon task plans, remains an important direction for future work. In addition, our method builds on a pretrained Janus-Flow VLM and performs reasoning--action alignment during downstream post-training; incorporating embodied CoT learning at pretraining stage may further improve generalization across tasks. Finally, by strengthening connection between embodied reasoning and actions, CoRAct provides a natural interface for agentic reasoning systems, where the broader knowledge and planning capabilities of agents could be grounded into more generalizable robotic behavior.
\section*{Acknowledgments}
We thank Tongze Liu, Kun Xu, Qingyang Gao, Haonan Liu,
Bin Zhang, Xinhua Wang, Pengcheng Li, and Li Wang
for their support of this work.
The completion of this work also relied on substantial
infrastructure support across the entire research pipeline.
We gratefully acknowledge the Beijing Innovation Center
of Humanoid Robotics for providing the robotic platforms,
experimental infrastructure, and operational support essential to this work.

\bibliographystyle{IEEEtran}
\bibliography{IEEEabrv,ref}

\begin{thebibliography}{10}
\providecommand{\url}[1]{#1}
\csname url@samestyle\endcsname
\providecommand{\newblock}{\relax}
\providecommand{\bibinfo}[2]{#2}
\providecommand{\BIBentrySTDinterwordspacing}{\spaceskip=0pt\relax}
\providecommand{\BIBentryALTinterwordstretchfactor}{4}
\providecommand{\BIBentryALTinterwordspacing}{\spaceskip=\fontdimen2\font plus
\BIBentryALTinterwordstretchfactor\fontdimen3\font minus \fontdimen4\font\relax}
\providecommand{\BIBforeignlanguage}[2]{{%
\expandafter\ifx\csname l@#1\endcsname\relax
\typeout{** WARNING: IEEEtran.bst: No hyphenation pattern has been}%
\typeout{** loaded for the language `#1'. Using the pattern for}%
\typeout{** the default language instead.}%
\else
\language=\csname l@#1\endcsname
\fi
#2}}
\providecommand{\BIBdecl}{\relax}
\BIBdecl

\bibitem{zawalski25a}
M.~Zawalski, W.~Chen, K.~Pertsch, O.~Mees, C.~Finn, and S.~Levine, ``Robotic control via embodied chain-of-thought reasoning,'' in \emph{Proceedings of The 8th Conference on Robot Learning}, ser. Proceedings of Machine Learning Research, P.~Agrawal, O.~Kroemer, and W.~Burgard, Eds., vol. 270.\hskip 1em plus 0.5em minus 0.4em\relax PMLR, 06--09 Nov 2025, pp. 3157--3181.

\bibitem{sunemma}
Q.~Sun, P.~Hong, T.~D. Pala, V.~Toh, U.-X. Tan, D.~Ghosal, and S.~Poria, ``Emma-{X}: An embodied multimodal action model with grounded chain of thought and look-ahead spatial reasoning,'' in \emph{Proceedings of the 63rd Annual Meeting of the Association for Computational Linguistics (Volume 1: Long Papers)}, W.~Che, J.~Nabende, E.~Shutova, and M.~T. Pilehvar, Eds.\hskip 1em plus 0.5em minus 0.4em\relax Vienna, Austria: Association for Computational Linguistics, Jul. 2025, pp. 14\,199--14\,214.

\bibitem{zhao2025cot}
Q.~Zhao, Y.~Lu, M.~J. Kim, Z.~Fu, Z.~Zhang, Y.~Wu, Z.~Li, Q.~Ma, S.~Han, C.~Finn \emph{et~al.}, ``Cot-vla: Visual chain-of-thought reasoning for vision-language-action models,'' in \emph{2025 IEEE/CVF Conference on Computer Vision and Pattern Recognition (CVPR)}.\hskip 1em plus 0.5em minus 0.4em\relax IEEE, 2025, pp. 1702--1713.

\bibitem{huang2025motvla}
W.~Huang, C.~Chen, H.~Qi, C.~Lv, Y.~Du, and H.~Yang, ``Motvla: A vision-language-action model with unified fast-slow reasoning,'' \emph{arXiv preprint arXiv:2510.18337}, 2025.

\bibitem{shou2026halo}
Q.~Shou, F.~Zhu, S.~Chen, P.~Yan, Z.~Yan, Y.~Miao, X.~Pang, Z.~Hong, R.~Shi, H.~Huang \emph{et~al.}, ``Halo: A unified vision-language-action model for embodied multimodal chain-of-thought reasoning,'' \emph{arXiv preprint arXiv:2602.21157}, 2026.

\bibitem{ma2026unifying}
X.~Ma, L.~Xing, H.~Zhang, W.~Li, and S.~Lu, ``Unifying perception and action: A hybrid-modality pipeline with implicit visual chain-of-thought for robotic action generation,'' in \emph{Proceedings of the IEEE/CVF Conference on Computer Vision and Pattern Recognition}, 2026, pp. 22\,380--22\,390.

\bibitem{lu2026thinkingvla}
T.~Lu, H.~Zhang, Z.~Diao, J.~Wang, S.~Xu, X.~Lin, G.~Zhong, Z.~Ye, P.~Wang, Z.~Wu \emph{et~al.}, ``Thinkingvla: Interleaved vision and language reasoning for robotic manipulation,'' \emph{arXiv preprint arXiv:2606.17937}, 2026.

\bibitem{zhang2026robogene}
\BIBentryALTinterwordspacing
Y.~Zhang, K.~Wu, Z.~Gao, Z.~Zhao, P.~Ren, Z.~Xu, F.~Liao, X.~Wang, S.~Fan, D.~Wu, Q.~Feng, M.~Li, Z.~Che, C.~Liu, and J.~Tang, ``Robogene: Diversity-driven agentic task generation framework for vla pre-training,'' in \emph{Conference on Robot Learning (CoRL)}, 2026, accepted at CoRL 2026. [Online]. Available: \url{https://arxiv.org/abs/2602.16444}
\BIBentrySTDinterwordspacing

\bibitem{wang2026roboaug}
\BIBentryALTinterwordspacing
X.~Wang, K.~Wu, Z.~Zhao, H.~Cao, Y.~Zhao, Z.~Xu, M.~Li, S.~Fan, D.~Wu, Y.~Zhang, N.~Liu, Z.~Che, and J.~Tang, ``Roboaug: One annotation to hundreds of scenes via region-contrastive data augmentation for robotic manipulation,'' in \emph{Conference on Robot Learning (CoRL)}, 2026, accepted at CoRL 2026. [Online]. Available: \url{https://arxiv.org/abs/2602.14032}
\BIBentrySTDinterwordspacing

\bibitem{fan2026xr1}
\BIBentryALTinterwordspacing
S.~Fan, K.~Wu, Z.~Che, X.~Wang, D.~Wu, F.~Liao, N.~Liu, Y.~Zhang, Z.~Zhao, Z.~Xu, M.~Li, Q.~Liu, S.~Zhang, M.~Wan, and J.~Tang, ``{XR}-1: Towards versatile vision-language-action models via learning unified vision-motion representations,'' in \emph{Proceedings of the 43rd International Conference on Machine Learning}, ser. Proceedings of Machine Learning Research, vol. 306.\hskip 1em plus 0.5em minus 0.4em\relax PMLR, 2026, pp. 28\,838--28\,882. [Online]. Available: \url{https://proceedings.mlr.press/v306/fan26f.html}
\BIBentrySTDinterwordspacing

\bibitem{kim2026roboalign}
D.~Kim, S.~Park, W.~Song, S.~Kim, T.~Kim, H.~Jang, J.~Shin, J.~Kim, and Y.~Seo, ``Roboalign: Learning test-time reasoning for language-action alignment in vision-language-action models,'' \emph{arXiv preprint arXiv:2603.21341}, 2026.

\bibitem{sun2026atomvla}
X.~Sun, Z.~Xu, C.~Cao, Z.~Liu, Y.~Sun, J.~Pang, R.~Zhang, Z.~Yang, K.~Pang, D.~He \emph{et~al.}, ``Atomvla: Scalable post-training for robotic manipulation via predictive latent world models,'' \emph{arXiv preprint arXiv:2603.08519}, 2026.

\bibitem{huang2026thinkact}
C.-P. Huang, Y.-H. Wu, M.-H. Chen, F.~Wang, and F.-E. Yang, ``Thinkact: Vision-language-action reasoning via reinforced visual latent planning,'' \emph{Advances in Neural Information Processing Systems}, vol.~38, pp. 82\,782--82\,802, 2026.

\bibitem{zhong2026acot}
L.~Zhong, Y.~Liu, Y.~Wei, Z.~Xiong, S.~Liu, and G.~Ren, ``Acot-vla: Action chain-of-thought for vision-language-action models,'' in \emph{Proceedings of the IEEE/CVF Conference on Computer Vision and Pattern Recognition}, 2026, pp. 8152--8162.

\bibitem{chen25a}
W.~Chen, S.~Belkhale, S.~Mirchandani, K.~Pertsch, D.~Driess, O.~Mees, and S.~Levine, ``Training strategies for efficient embodied reasoning,'' in \emph{Proceedings of The 9th Conference on Robot Learning}, ser. Proceedings of Machine Learning Research, J.~Lim, S.~Song, and H.-W. Park, Eds., vol. 305.\hskip 1em plus 0.5em minus 0.4em\relax PMLR, 27--30 Sep 2025, pp. 365--391.

\bibitem{bai2026latent}
S.~Bai, J.~Lyu, W.~Zhou, Z.~Li, D.~Wang, L.~Xing, X.~Zhao, P.~Wang, Z.~Wang, C.~Chi \emph{et~al.}, ``Latent reasoning vla: Latent thinking and prediction for vision-language-action models,'' \emph{arXiv preprint arXiv:2602.01166}, 2026.

\bibitem{liu2026last}
Z.~Liu, J.~Liu, H.~Chen, J.~Yu, Z.~Guo, C.~Hou, C.~Gu, X.~Mi, R.~Zhang, K.~Wu \emph{et~al.}, ``Last $ \_ $\{$0$\}$ $: Latent spatio-temporal chain-of-thought for robotic vision-language-action model,'' \emph{arXiv preprint arXiv:2601.05248}, 2026.

\bibitem{sun2026revisiting}
N.~Sun, Y.~Zhang, Y.~Yang, W.~Zhao, P.~Li, J.~Guo, W.~Song, P.~Ding, R.~Suo, Y.~Su \emph{et~al.}, ``Revisiting embodied chain-of-thought for generalizable robot manipulation,'' \emph{arXiv preprint arXiv:2606.03784}, 2026.

\bibitem{Wulff_2026_CVPR}
T.~Wulff, F.~Tavella, R.~S. Maharjan, M.~Adikari, and A.~Cangelosi, ``Grounding hierarchical vision-language-action models through explicit language-action alignment,'' in \emph{Proceedings of the IEEE/CVF Conference on Computer Vision and Pattern Recognition (CVPR) Findings}, June 2026, pp. 9269--9281.

\bibitem{ma2026dvlm}
Y.~Ma, Y.~Cao, W.~Ding, S.~Zhang, Y.~Wang, B.~Ivanovic, M.~Jiang, M.~Pavone, and C.~Xiao, ``dvlm-ad: Enhance diffusion vision-language-model for driving via controllable reasoning,'' in \emph{Proceedings of the IEEE/CVF Conference on Computer Vision and Pattern Recognition}, 2026, pp. 1050--1061.

\bibitem{tan2025latent}
S.~Tan, K.~Chitta, Y.~Chen, R.~Tian, Y.~You, Y.~Wang, W.~Luo, Y.~Cao, P.~Krahenbuhl, M.~Pavone \emph{et~al.}, ``Latent chain-of-thought world modeling for end-to-end driving,'' \emph{arXiv preprint arXiv:2512.10226}, 2025.

\bibitem{wu2025foresight}
Y.~Wu, R.~Tian, G.~Swamy, and A.~Bajcsy, ``From foresight to forethought: Vlm-in-the-loop policy steering via latent alignment,'' \emph{arXiv preprint arXiv:2502.01828}, 2025.

\bibitem{xia2025apo}
W.~Xia, Y.~Yang, H.~Wu, X.~Ma, T.~Kong, and D.~Hu, ``Human-assisted robotic policy refinement via action preference optimization,'' in \emph{Advances in Neural Information Processing Systems}, vol.~38, 2025.

\bibitem{xia2026robovaluerl}
\BIBentryALTinterwordspacing
W.~Xia, P.~Ren, W.~Yu, Y.~Zhang, J.~Li, Y.~Zhang, Y.~Zhao, Q.~Gao, J.~Fu, J.~Tang, J.-R. Wen, Z.~Che, and D.~Hu, ``Robo-valuerl: Reliable value estimation for offline-to-online reinforcement learning,'' in \emph{Conference on Robot Learning (CoRL)}, 2026, accepted at CoRL 2026. [Online]. Available: \url{https://arxiv.org/abs/2607.09866}
\BIBentrySTDinterwordspacing

\bibitem{xia2025phoenix}
W.~Xia, R.~Feng, D.~Wang, and D.~Hu, ``Phoenix: A motion-based self-reflection framework for fine-grained robotic action correction,'' in \emph{Proceedings of the IEEE/CVF Conference on Computer Vision and Pattern Recognition}, 2025, pp. 6981--6990.

\bibitem{kwok2026scaling}
J.~Kwok, X.~Zhang, M.~Xu, Y.~Liu, A.~Mirhoseini, C.~Finn, and M.~Pavone, ``Scaling verification can be more effective than scaling policy learning for vision-language-action alignment,'' \emph{arXiv preprint arXiv:2602.12281}, 2026.

\bibitem{wu2025you}
Y.~Wu, A.~Li, T.~Hermans, F.~Ramos, A.~Bajcsy, and C.~P{\'e}rez-D'Arpino, ``Do what you say: Steering vision-language-action models via runtime reasoning-action alignment verification,'' \emph{arXiv preprint arXiv:2510.16281}, 2025.

\bibitem{kwok25a}
J.~Kwok, C.~Agia, R.~Sinha, M.~Foutter, S.~Li, I.~Stoica, A.~Mirhoseini, and M.~Pavone, ``Robomonkey: Scaling test-time sampling and verification for vision-language-action models,'' in \emph{Proceedings of The 9th Conference on Robot Learning}, ser. Proceedings of Machine Learning Research, J.~Lim, S.~Song, and H.-W. Park, Eds., vol. 305.\hskip 1em plus 0.5em minus 0.4em\relax PMLR, 27--30 Sep 2025, pp. 3200--3217.

\bibitem{dai2025rover}
M.~Dai, L.~Liu, Y.~Bai, Y.~Liu, Z.~Wang, R.~Su, C.~Chen, L.~Lin, and X.~Wu, ``Rover: Robot reward model as test-time verifier for vision-language-action model,'' \emph{arXiv preprint arXiv:2510.10975}, 2025.

\bibitem{ye2025vla}
A.~Ye, Z.~Zhang, B.~Wang, X.~Wang, D.~Zhang, and Z.~Zhu, ``Vla-r1: Enhancing reasoning in vision-language-action models,'' \emph{arXiv preprint arXiv:2510.01623}, 2025.

\bibitem{yin2025deepthinkvla}
C.~Yin, Y.~Lin, W.~Xu, S.~Tam, X.~Zeng, Z.~Liu, and Z.~Yin, ``Deepthinkvla: Enhancing reasoning capability of vision-language-action models,'' \emph{arXiv preprint arXiv:2511.15669}, 2025.

\bibitem{wang2025alpamayo}
Y.~Wang, W.~Luo, J.~Bai, Y.~Cao, T.~Che, K.~Chen, Y.~Chen, J.~Diamond, Y.~Ding, W.~Ding \emph{et~al.}, ``Alpamayo-r1: Bridging reasoning and action prediction for generalizable autonomous driving in the long tail,'' \emph{arXiv preprint arXiv:2511.00088}, 2025.

\bibitem{foutter2026vision}
M.~Foutter, M.~Cercola, L.~Wild, Y.~Wang, M.~Li, D.~Gammelli, and M.~Pavone, ``Do vision-language-action models mean what they say? on the role of faithfulness in embodied reasoning,'' \emph{arXiv preprint arXiv:2607.04681}, 2026.

\bibitem{liang2024mixture}
W.~Liang, L.~Yu, L.~Luo, S.~Iyer, N.~Dong, C.~Zhou, G.~Ghosh, M.~Lewis, W.-t. Yih, L.~Zettlemoyer \emph{et~al.}, ``Mixture-of-transformers: A sparse and scalable architecture for multi-modal foundation models,'' \emph{arXiv preprint arXiv:2411.04996}, 2024.

\bibitem{li2025gr}
Y.~Li, X.~Ma, J.~Xu, Y.~Cui, Z.~Cui, Z.~Han, L.~Huang, T.~Kong, Y.~Liu, H.~Niu \emph{et~al.}, ``Gr-rl: Going dexterous and precise for long-horizon robotic manipulation,'' \emph{arXiv preprint arXiv:2512.01801}, 2025.

\bibitem{wagenmaker2025steering}
A.~Wagenmaker, M.~Nakamoto, Y.~Zhang, S.~Park, W.~Yagoub, A.~Nagabandi, A.~Gupta, and S.~Levine, ``Steering your diffusion policy with latent space reinforcement learning,'' \emph{arXiv preprint arXiv:2506.15799}, 2025.

\bibitem{shao2024deepseekmath}
Z.~Shao, P.~Wang, Q.~Zhu, R.~Xu, J.~Song, X.~Bi, H.~Zhang, M.~Zhang, Y.~Li, Y.~Wu \emph{et~al.}, ``Deepseekmath: Pushing the limits of mathematical reasoning in open language models,'' \emph{arXiv preprint arXiv:2402.03300}, 2024.

\bibitem{ma2025janusflow}
Y.~Ma, X.~Liu, X.~Chen, W.~Liu, C.~Wu, Z.~Wu, Z.~Pan, Z.~Xie, H.~Zhang, X.~Yu \emph{et~al.}, ``Janusflow: Harmonizing autoregression and rectified flow for unified multimodal understanding and generation,'' in \emph{2025 IEEE/CVF Conference on Computer Vision and Pattern Recognition (CVPR)}.\hskip 1em plus 0.5em minus 0.4em\relax IEEE, 2025, pp. 7739--7751.

\bibitem{podell2024sdxl}
D.~Podell, Z.~English, K.~Lacey, A.~Blattmann, T.~Dockhorn, J.~M{\"u}ller, J.~Penna, and R.~Rombach, ``Sdxl: Improving latent diffusion models for high-resolution image synthesis,'' in \emph{International Conference on Learning Representations}, vol. 2024, 2024, pp. 1862--1874.

\bibitem{nasiriany2024robocasa}
S.~Nasiriany, A.~Maddukuri, L.~Zhang, A.~Parikh, A.~Lo, A.~Joshi, A.~Mandlekar, and Y.~Zhu, ``Robocasa: Large-scale simulation of everyday tasks for generalist robots,'' \emph{arXiv preprint arXiv:2406.02523}, 2024.

\bibitem{zhang2025vlabench}
S.~Zhang, Z.~Xu, P.~Liu, X.~Yu, Y.~Li, Q.~Gao, Z.~Fei, Z.~Yin, Z.~Wu, Y.-G. Jiang \emph{et~al.}, ``Vlabench: A large-scale benchmark for language-conditioned robotics manipulation with long-horizon reasoning tasks,'' in \emph{2025 IEEE/CVF International Conference on Computer Vision (ICCV)}.\hskip 1em plus 0.5em minus 0.4em\relax IEEE, 2025, pp. 11\,142--11\,152.

\end{thebibliography}

\end{document}